\documentclass[10pt,twocolumn,letterpaper]{article}

\usepackage{cvpr}              

\usepackage{amssymb}
\usepackage{booktabs}  
\usepackage{graphicx}  
\usepackage{pifont}

\definecolor{cvprblue}{rgb}{0.21,0.49,0.74}
\usepackage[pagebackref,breaklinks,colorlinks,allcolors=cvprblue]{hyperref}

\def\paperID{6039} 
\def\confName{CVPR}
\def\confYear{2026}

\title{Lang2Motion: Bridging Language and Motion through Joint Embedding Spaces}

\author{Bishoy Galoaa, Xiangyu Bai, Sarah Ostadabbas\\
Northeastern University\\
Boston, MA, USA\\
{\tt\small \{galoaa.b, bai.xiang, s.ostadabbas\}@northeastern.edu}
}
\begin{document}

\twocolumn[{
\renewcommand\twocolumn[1][]{#1}
\maketitle
\begin{center}
    \vspace{-28pt}
    \includegraphics[width=1.0\linewidth]{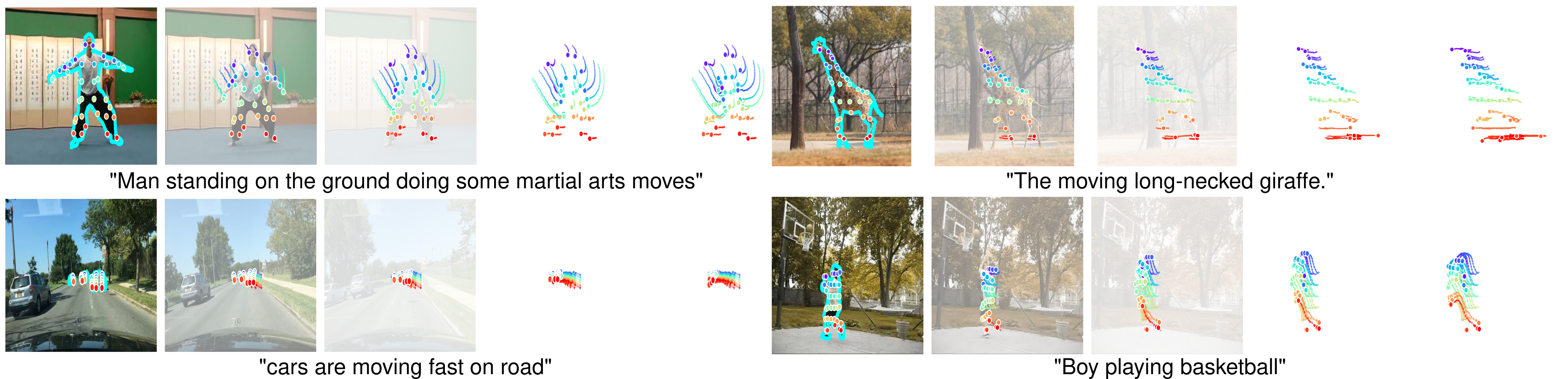}
    \centering
    \captionsetup{type=figure}
    \vspace{-5pt}
   \caption{Lang2Motion generates point trajectories from natural language descriptions via CLIP's joint embedding space. Each row shows the progression from video context (left) to pure trajectory generation (right), with fading frames illustrating our approach: training on trajectories extracted from real videos via point tracking, but generating motion without video synthesis. Starting from point grids (cyan), our method produces physically valid motion across diverse scenarios: (top-left) \textit{``Man standing on the ground doing some martial arts moves''} with coherent full-body dynamics, (top-right) \textit{``The moving long-necked giraffe''} maintaining trajectory consistency through tree occlusion, (bottom-left) \textit{``cars are moving fast on road''} where distant small objects exhibit correct motion, and (bottom-right) \textit{``Boy playing basketball''} with realistic motion transitions. The visualization demonstrates how we bridge video-based learning with video-free generation by extracting motion understanding from real footage to generate explicit trajectory coordinates from text alone.}
    \label{fig:teaser}
\end{center}
}]

\maketitle


\newcommand{\TrajCLIPjes}{
\begin{figure*}[t]
    \centering
    \includegraphics[width=\linewidth]{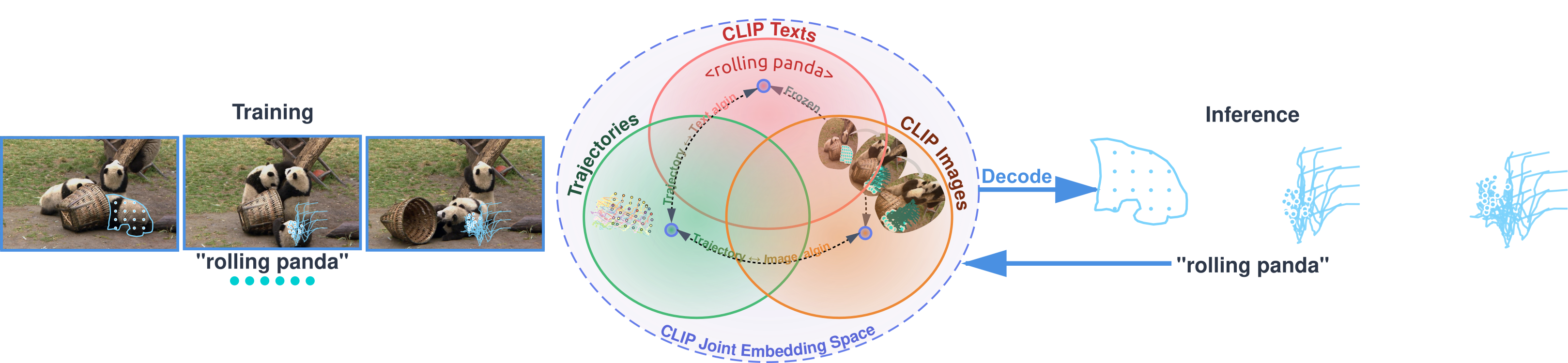}
    \caption{\textbf{Lang2Motion Joint Embedding Space Alignment.} During training (left), real-world videos with motion (e.g., \textit{``rolling panda''}) and extracted \textcolor{cyan}{point trajectories} from initial grids are used to align three modalities in CLIP's joint embedding space. The center shows how \textcolor{red}{text descriptions}, \textcolor{green}{trajectory representations}, and \textcolor{orange}{visual motion patterns} converge through alignment, connecting language semantics with geometric motion and visual appearance. At inference (right), a text description and initial grid mask decode into physically valid trajectory sequences, generating motion without video synthesis.}
    \vspace{-0.5cm}
    \label{fig:joint_embedding}
\end{figure*}

}
\newcommand{\TrajCLIPmethod}{
\begin{figure*}[t]
    \centering
    \includegraphics[width=0.80\linewidth]{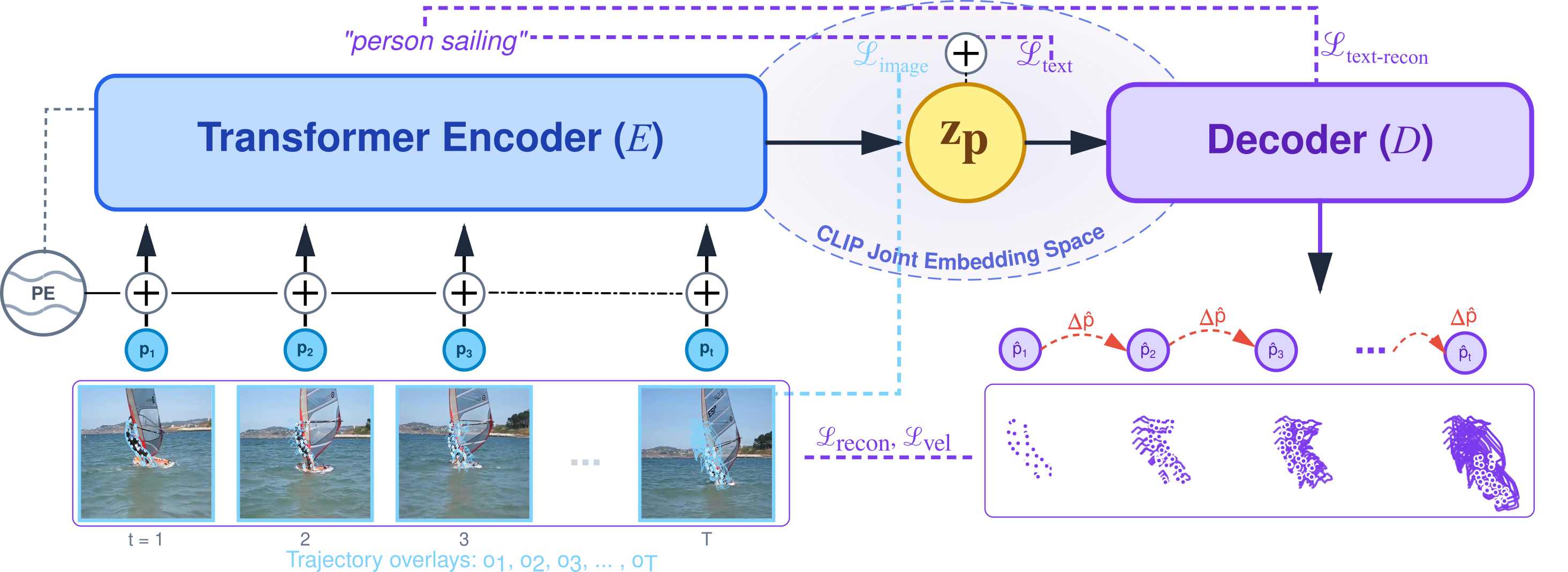}
    \caption{\textbf{Lang2Motion Architecture.} Point trajectories $\mathbf{p}_{1:T}$ {\color{cyan}(cyan)} with trajectory overlays $\mathbf{o}_{1:T}$ rendered on video frames are enriched with positional embeddings and a learned prefix token $\mathbf{z}_{tk}$, then encoded via transformer encoder $E$ into latent representations $\mathbf{z}_p$ in CLIP's joint embedding space. Dual alignment losses connect $\mathbf{z}_p$ with CLIP text embeddings via {\color{violet}$\mathcal{L}_{\text{text}}$} and CLIP image embeddings of trajectory overlays via {\color{blue}$\mathcal{L}_{\text{image}}$}. The decoder $D$ autoregressively reconstructs trajectory sequences $\hat{\mathbf{p}}_{1:T}$ {\color{violet}(violet)} by generating frame displacements {\color{red}$\Delta\mathbf{p}_t$} conditioned on $\mathbf{z}_p$ and previous positions $\mathbf{p}_{t-1}^*$, supervised by {\color{violet}$\mathcal{L}_{\text{recon}}$} and {\color{violet}$\mathcal{L}_{\text{vel}}$}. {\color{violet}$\mathcal{L}_{\text{text-recon}}$} enables direct text-to-trajectory generation by training $D$ to decode from $\text{CLIP}_{\text{text}}(t)$. At inference, text descriptions are encoded via CLIP's frozen text encoder and decoded directly to generate trajectory sequences.}
    \label{fig:architecture}
    \vspace{-0.5cm}
\end{figure*}
}

\newcommand{\langmotioninterpolate}{
\begin{figure*}[t]
    \centering
    \includegraphics[width=\linewidth]{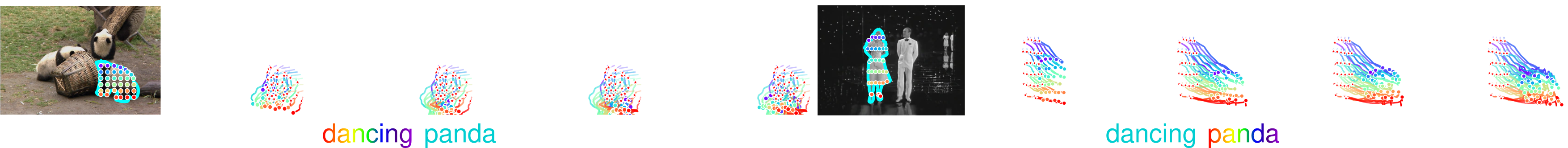}
    \caption{\textbf{Trajectory Generation from Initial Grid and Latent Space Interpolation.} Given identical text \textit{``dancing panda''}, Lang2Motion generates different motion interpretations based on initial grid placement. \textbf{Left:} From a panda's grid, the model emphasizes dancing motion: {\color{orange}d}{\color{yellow}a}{\color{green}n}{\color{cyan}c}{\color{blue}i}{\color{purple}n}{\color{magenta}g} {\color{cyan}panda}. \textbf{Right:} From a dancing pose grid, the model emphasizes panda appearance: {\color{cyan}dancing} {\color{orange}p}{\color{yellow}a}{\color{green}n}{\color{cyan}d}{\color{blue}a}. Initial grids use automatically retrieved masks; initial video frames shown for visualization only. This demonstrates semantic interpolation in CLIP's joint embedding space. See supplementary material for additional results.}
    \label{fig:interpolation}
\end{figure*}
}

\newcommand{\motiongeneval}{
\begin{figure}[t]
    \centering
    \includegraphics[width=\columnwidth]{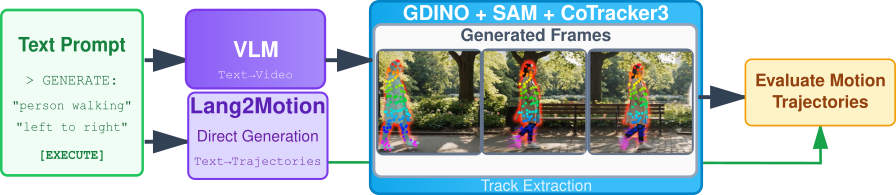}
    \caption{\textbf{Video-Based VLM Evaluation Pipeline.} Comparison of VLM-based and Lang2Motion trajectory generation. VLM generates video frames from which trajectories are extracted using GDINO~\cite{liu2024grounding}+SAM~\cite{kirillov2023segment}+CoTracker3~\cite{karaev2024cotracker3}, while Lang2Motion generates trajectories directly. Both outputs are evaluated using identical motion quality metrics.}
    \label{fig:vlm_eval}
\end{figure}
}

\newcommand{\motiongencomp}{
\begin{figure}[t]
    \centering
    \includegraphics[width=\columnwidth]{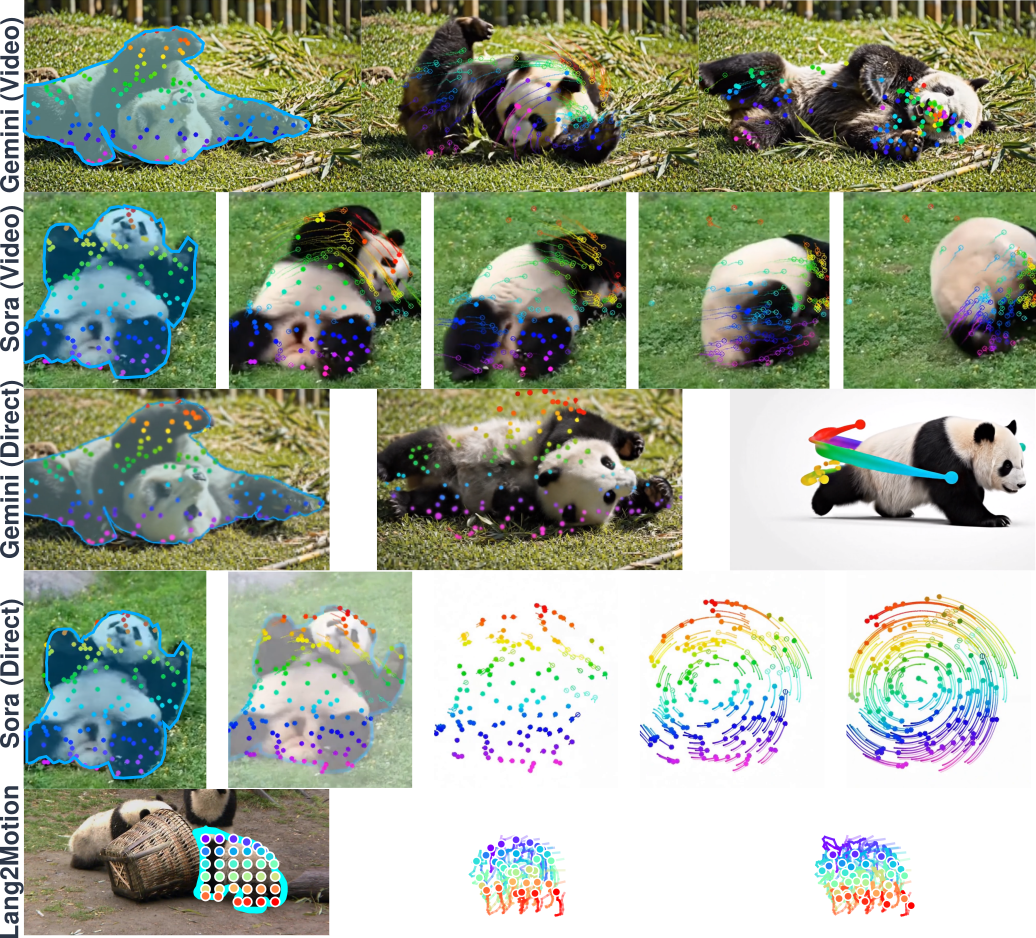}
    \caption{\textbf{Qualitative Comparison.} Video-based methods (Rows 1-2: Gemini Veo3, Sora 2.0) generate frames requiring trajectory extraction. Direct VLM generation (Rows 3-4) produces incoherent motion (Gemini) or loses object structure despite correct motion semantics (Sora). Lang2Motion (Row 5) generates physically valid trajectories from text and initial mask, maintaining both motion semantics (rolling) and object coherence (panda) throughout.}
    \label{fig:vlm_com}
\end{figure}
}
\newcommand{\motionclipcomparison}{
\begin{figure}[t]
\centering
\includegraphics[width=\columnwidth]{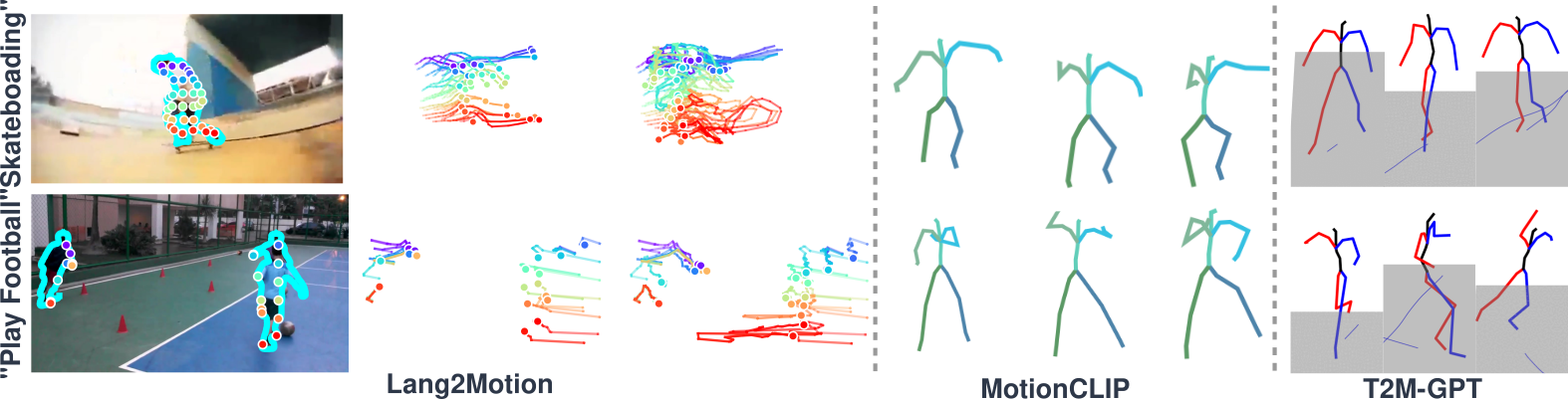}
\caption{\textbf{Comparison with Human Motion Generation Methods.} Lang2Motion generates temporally-aware trajectories capturing real-world motion dynamics, while MotionCLIP~\cite{tevet2022motionclip} produces anatomically correct poses without continuous temporal flow. For ``Play Football'', MotionCLIP generates a Ronaldo celebration instead of playing motion, while Lang2Motion produces coordinated multi-subject trajectories. T2M-GPT~\cite{zhang2023t2m} produces wrong pose and does not supportmulti-object scenes.}
\vspace{-0.5cm}
\label{fig:motionclip_comparison}
\end{figure}
}
\newcommand{\retrievaltable}{
\begin{table}[t]
\centering
\caption{\textbf{Text-to-Trajectory Retrieval.} Comparison using Recall@K metrics. Video-based methods extract trajectories from videos, while Lang2Motion operates directly on trajectory representations. Random baseline for avg 15 expressions per video).}
\label{tab:retrieval}
\resizebox{\columnwidth}{!}{
\begin{tabular}{l|c|cccc}
\toprule
\textbf{Method} & \textbf{Type} & \textbf{R@1↑} & \textbf{R@3↑} & \textbf{R@5↑} & \textbf{R@10↑} \\
\midrule
Random & Baseline & 0.3 & 0.9 & 1.5 & 2.9 \\
CLIP NN & Retrieval & 12.3 & 28.4 & 39.7 & 54.2 \\
\midrule
CLIP4Clip~\cite{luo2021clip4clip} & Video-based & 18.5 & 38.2 & 51.4 & 68.3 \\
X-CLIP~\cite{ma2022x} & Video-based & 21.7 & 42.6 & 55.8 & 72.1 \\
\midrule
\textbf{Lang2Motion (Ours)} & Trajectory & \textbf{34.2} & \textbf{58.7} & \textbf{71.3} & \textbf{84.6} \\
\bottomrule
\end{tabular}
}
\end{table}
}
\newcommand{\generationtable}{
\begin{table}[t]
\centering
\caption{\textbf{Trajectory Generation Quality \& Efficiency.} We evaluate using TRAJAN motion metrics~\cite{allen2025direct} (AJ: Average Jaccard, OA: Occlusion Accuracy), trajectory accuracy (ADE, FDE), semantic alignment (CLIP sim), and speed. Video-based methods generate full videos then extract trajectories; Direct methods prompt VLMs to output coordinate sequences as text; Lang2Motion generates trajectories from text and initial masks.}
\label{tab:generation}
\resizebox{\columnwidth}{!}{
\begin{tabular}{l|cc|cc|cc|c}
\toprule
\textbf{Method} & \textbf{AJ↑} & \textbf{OA↑} & \textbf{ADE↓} & \textbf{FDE↓} & \textbf{CLIP sim↑} & \textbf{Smooth↑} & \textbf{Speed} \\
\midrule
\multicolumn{8}{c}{\textit{Video-Based Generation (with trajectory extraction)}} \\
\midrule
Gemini Veo3~\cite{deepmind2024veo} & 0.42 $\pm$ 0.03 & 0.88 $\pm$ 0.03 & 18.3 & 32.1 & 0.68 & 0.72 & 90s \\
Sora 2.0~\cite{brookssim} & 0.45 $\pm$ 0.02 & 0.92 $\pm$ 0.02 & 19.1 & 33.8 & 0.68 & 0.72 & 90s \\
\midrule
\multicolumn{8}{c}{\textit{Direct Trajectory Generation (VLM text output)}} \\
\midrule
Gemini Veo3 Direct & 0.28 $\pm$ 0.04 & 0.76 $\pm$ 0.04 & 24.6 & 41.2 & 0.54 & 0.58 & 8s \\
Sora 2.0 Direct & 0.31 $\pm$ 0.03 & 0.79 $\pm$ 0.03 & 25.3 & 42.7 & 0.54 & 0.58 & 8s \\
\midrule
\textbf{Lang2Motion (Ours)} & \textbf{0.84 $\pm$ 0.02} & \textbf{0.98 $\pm$ 0.01} & \textbf{12.4} & \textbf{21.7} & \textbf{0.81} & \textbf{0.89} & \textbf{0.7s} \\
\bottomrule
\end{tabular}
}
\end{table}
}
\newcommand{\ablationtable}{
\begin{table}[t]
\centering
\caption{\textbf{Ablation Study: Loss Components.} Impact of loss components on retrieval accuracy (R@1), semantic alignment (CLIP Sim), and reconstruction error (ADE). All configurations use identical model architecture and decoding strategy (autoregressive).}
\label{tab:ablation}
\resizebox{\columnwidth}{!}{
\begin{tabular}{l|ccc|c}
\toprule
\textbf{Configuration} & \textbf{R@1↑} & \textbf{CLIP Sim↑} & \textbf{ADE↓} & \textbf{Speed} \\
\midrule
Lang2Motion (full) & \textbf{34.2} & \textbf{0.81} & \textbf{12.4} & \textbf{0.7s} \\
\midrule
w/o $\mathcal{L}_{\text{text}}$ & 22.6 & 0.63 & 14.8 & 0.7s \\
w/o $\mathcal{L}_{\text{image}}$ & 19.8 & 0.59 & 15.2 & 0.7s \\
w/o $\mathcal{L}_{\text{text-recon}}$ & 28.3 & 0.74 & 18.6 & 0.7s \\
w/o both CLIP losses & 8.4 & 0.41 & 22.1 & 0.7s \\
\bottomrule
\end{tabular}
}
\end{table}
}

\newcommand{\reconstructiondecodingablation}{
\begin{table}[t]
\centering
\caption{\textbf{Ablation Study: Reconstruction Loss \& Decoding Strategy.} Comparison of reconstruction loss functions and decoding strategies.}
\label{tab:recon_decoding_ablation}
\resizebox{\columnwidth}{!}{
\begin{tabular}{l|cccc|c}
\toprule
\textbf{Configuration} & \textbf{ADE↓} & \textbf{FDE↓} & \textbf{Smooth↑} & \textbf{CLIP Sim↑} & \textbf{Speed} \\
\midrule
\multicolumn{6}{c}{\textit{Reconstruction Loss Function (Autoregressive Decoding)}} \\
\midrule
MSE (L2) & 16.8 & 28.3 & 0.83 & 0.76 & 0.7s \\
MAE (L1) & \textbf{12.4} & \textbf{21.7} & \textbf{0.89} & \textbf{0.81} & 0.7s \\
\midrule
\multicolumn{6}{c}{\textit{Decoding Strategy (L1 Reconstruction Loss)}} \\
\midrule
Direct (all frames) & 15.7 & 26.3 & 0.81 & 0.79 & 0.3s \\
Autoregressive & \textbf{12.4} & \textbf{21.7} & \textbf{0.89} & \textbf{0.81} & 0.7s \\
Teacher Forcing (train only) & 11.8 & 20.9 & 0.91 & 0.82 & --- \\
\bottomrule
\end{tabular}
}
\end{table}
}


\newcommand{\overlayablationtable}{
\begin{table}[t]
\centering
\caption{\textbf{Ablation Study: Trajectory Overlay Visualization.} Impact of trajectory color and opacity on CLIP alignment.}
\label{tab:overlay_ablation}
\resizebox{\columnwidth}{!}{
\begin{tabular}{l|ccccccc|cccc}
\toprule
& \multicolumn{7}{c|}{\textbf{Color Ablation (Opacity = 0.5)}} & \multicolumn{4}{c}{\textbf{Opacity Ablation (Cyan)}} \\
\midrule
\textbf{Color} & \textcolor{cyan}{Cyan} & \textcolor{blue}{Blue} & \colorbox{black}{\textcolor{white}{White}} & \textcolor{red}{Red} & \textcolor{green}{Green} & \textcolor{yellow}{Yellow} & Black & \textcolor{cyan!30}{Cyan} & \textcolor{cyan!50}{Cyan} & \textcolor{cyan!70}{Cyan} & \textcolor{cyan}{Cyan} \\
\textbf{Opacity} & 0.5 & 0.5 & 0.5 & 0.5 & 0.5 & 0.5 & 0.5 & 0.3 & 0.5 & 0.7 & 1.0 \\
\midrule
\textbf{CLIP Sim↑} & \textbf{0.81} & 0.76 & 0.74 & 0.73 & 0.71 & 0.69 & 0.66 & 0.75 & \textbf{0.81} & 0.78 & 0.72 \\
\textbf{R@1↑} & \textbf{34.2} & 30.5 & 29.1 & 28.4 & 26.9 & 25.7 & 23.8 & 30.8 & \textbf{34.2} & 32.6 & 28.9 \\
\bottomrule
\end{tabular}
}
\end{table}
}
\newcommand{\actionrecognitiontable}{
\begin{table}[t]
\centering
\caption{\textbf{Human Action Recognition from Pose Trajectories.} Comparison on standard action recognition benchmarks. Lang2Motion operates on skeletal pose trajectories, demonstrating that our trajectory-aligned embedding space generalizes to structured human motion.}
\label{tab:action_recognition}
\resizebox{\columnwidth}{!}{
\begin{tabular}{l|c|cc}
\toprule
\textbf{Method} & \textbf{Dataset} & \textbf{Top-1 Acc↑} & \textbf{Top-5 Acc↑} \\
\midrule
ST-GCN~\cite{yan2018spatial} & NTU RGB+D~\cite{shahroudy2016ntu} & 81.5 & 94.2 \\
MotionCLIP~\cite{tevet2022motionclip} & NTU RGB+D~\cite{shahroudy2016ntu} & 76.7 & 91.3 \\
MS-G3D~\cite{liu2020ms} & NTU RGB+D~\cite{shahroudy2016ntu} & 86.9 & 96.8 \\
\textbf{Lang2Motion (Ours)} & NTU RGB+D~\cite{shahroudy2016ntu} & \textbf{88.3} & \textbf{97.2} \\
\midrule
ST-GCN~\cite{yan2018spatial} & Kinetics-Skeleton~\cite{kay2017kinetics} & 30.7 & 52.8 \\
MotionCLIP~\cite{tevet2022motionclip} & Kinetics-Skeleton~\cite{kay2017kinetics} & 35.2 & 58.4 \\
MS-G3D~\cite{liu2020ms} & Kinetics-Skeleton~\cite{kay2017kinetics} & 38.0 & 60.9 \\
\textbf{Lang2Motion (Ours)} & Kinetics-Skeleton~\cite{kay2017kinetics} & \textbf{41.6} & \textbf{64.3} \\
\bottomrule
\end{tabular}
}
\end{table}
}


\newcommand{\queryfreetable}{
\begin{table}[t]
\centering
\caption{\textbf{Multi-Modal Trajectory Generation.} Lang2Motion generates trajectories from multiple input modalities, demonstrating CLIP-aligned space flexibility. Text provides semantic control (baseline), while image+mask leverages visual understanding without explicit descriptions.}
\label{tab:query_free}
\resizebox{0.85\columnwidth}{!}{
\begin{tabular}{l|ccc}
\toprule
\textbf{Input Modality} & \textbf{CLIP Sim↑} & \textbf{ADE↓} & \textbf{Smoothness↑} \\
\midrule
Text + Mask (baseline) & \textbf{0.81} & \textbf{12.4} & \textbf{0.89} \\
\midrule
Image + Mask (no text) & 0.76 & 14.2 & 0.85 \\
Trajectory Overlay Only & 0.72 & 15.8 & 0.82 \\
\bottomrule
\end{tabular}
}
\end{table}
}

\newcommand{\gridablationtable}{
\begin{table}[t]
\centering
\caption{\textbf{Ablation Study: Grid Size.} Impact of point grid configuration on trajectory quality and semantic alignment. Temporal length fixed at $T=30$ frames.}
\label{tab:grid_ablation}
\resizebox{0.85\columnwidth}{!}{
\begin{tabular}{l|c|cccc}
\toprule
\textbf{Grid Size} & \textbf{Points (N)} & \textbf{R@1↑} & \textbf{CLIP Sim↑} & \textbf{ADE↓} & \textbf{FDE↓} \\
\midrule
$3 \times 3$ & 9 & 28.4 & 0.73 & 16.8 & 29.2 \\
$4 \times 4$ & 16 & 31.2 & 0.77 & 14.1 & 24.8 \\
$5 \times 5$ & 25 & 33.1 & 0.79 & 13.2 & 23.1 \\
$6 \times 6$ & \textbf{36} & \textbf{34.2} & \textbf{0.81} & \textbf{12.4} & \textbf{21.7} \\
$8 \times 8$ & 64 & 33.8 & 0.80 & 12.6 & 22.1 \\
$10 \times 10$ & 100 & 33.5 & 0.79 & 12.9 & 22.8 \\
\bottomrule
\end{tabular}
}
\end{table}
}

\begin{abstract}
We present Lang2Motion, a framework for language-guided point trajectory generation by aligning motion manifolds with joint embedding spaces. Unlike prior work focusing on human motion or video synthesis, we generate explicit trajectories for arbitrary objects using motion extracted from real-world videos via point tracking. Our transformer-based auto-encoder learns trajectory representations through dual supervision: textual motion descriptions and rendered trajectory visualizations, both mapped through CLIP's frozen encoders. Lang2Motion achieves 34.2\% Recall@1 on text-to-trajectory retrieval, outperforming video-based methods by 12.5 points, and improves motion accuracy by 33-52\% (12.4 ADE vs 18.3-25.3) compared to video generation baselines. We demonstrate 88.3\% Top-1 accuracy on human action recognition despite training only on diverse object motions, showing effective transfer across motion domains. Lang2Motion supports style transfer, semantic interpolation, and latent-space editing through CLIP-aligned trajectory representations. Code is available at \url{https://github.com/ostadabbas/Lang2Motion}.
\end{abstract}    
\section{Introduction}
\label{sec:intro}
Imagine describing motion with words: \textit{``a person performing martial arts,''}, \textit{``a giraffe walking through trees,''},  \textit{``a basketball player jumping to shoot''}, and having a computer generate the precise trajectories these descriptions evoke (Figure~\ref{fig:teaser}). While humans effortlessly translate language into motion patterns, bridging the semantic richness of natural language with the geometric precision of point trajectories remains a fundamental challenge. This gap limits intuitive control over motion in animation, robotics, video editing, and physical simulation. Lang2Motion (Language to Motion) generates point trajectories through language-guided control by aligning real-world motion with image-text joint embedding space (Figure~\ref{fig:joint_embedding}).
\TrajCLIPjes

Joint embedding spaces can bridge disparate modalities through aligned semantic representations. CLIP~\cite{radford2021learning}, trained on image-text pairs, created a space where visual and linguistic concepts naturally align. This paradigm extends beyond static images: AudioCLIP~\cite{guzhov2021audioclip} to audio, MotionCLIP~\cite{tevet2022motionclip} to human skeletal motion, and ImageBind~\cite{girdhar2023imagebind} to multiple modalities. These successes share a common insight: semantic structure learned from one domain transfers to others through careful alignment, without retraining the original embedding space.

Yet arbitrary point trajectories from real-world videos remain unexplored. Existing work aligns human motion~\cite{tevet2022motionclip,tevet2022human}, video sequences~\cite{singer2022make,assran2025vjepa2}, or 3D geometry~\cite{poole2022dreamfusion} with semantic spaces, but none generate explicit point trajectories from text. This gap matters because trajectories represent the fundamental language of motion; capturing \textit{how} objects move spatiotemporally, applicable to any moving entity from particles to animals to deformable objects.

Lang2Motion bridges semantic representation and physical motion understanding by aligning point-trajectory manifolds with vision–language embedding spaces. We leverage dense point tracking to extract fine-grained trajectories from MeViS~\cite{ding2023mevis}, capturing natural motion phenomena (gravity, momentum, collision, deformation) from real videos. These trajectories are projected into a shared embedding space via dual supervision: (1) textual motion descriptions (e.g., ``person walking left,'' ``bird flying upward'') and (2) rendered trajectory visualizations, both encoded through frozen vision–language encoders. Once aligned, Lang2Motion exhibits emergent motion extrapolation capabilities, continuing plausible trajectories without continuous visual input, akin to human motion inference during occlusion or brief visual absence. This phenomenon is illustrated in Figure~\ref{fig:teaser}, where the model preserves a giraffe's motion trajectory through tree occlusion. This behavior parallels human perception of biological motion kinematograms, where observers effortlessly recognize actions and object identities from sparse point-light displays~\cite{saleki2023objects}. In such displays, motion cues alone convey sufficient information for recognition, even when texture and contour are absent. Similarly, at inference, Lang2Motion uses text embeddings to control trajectory generation, with point grids initialized from retrieved or specified object masks, generating realistic motion paths drawn from the aligned trajectory manifold, demonstrating that motion semantics can emerge naturally from language-grounded alignment.

Critically, the trajectories used in Lang2Motion are real-world motion sequences extracted from videos rather than synthetic constructs. Leveraging CoTracker3~\cite{karaev2024cotracker3}, we obtain dense point trajectories from MeViS~\cite{ding2023mevis} videos by initializing a 6×6 grid of points within object segmentation masks, achieving over higher data efficiency compared to frame-based modeling. Points initialized within object masks capture natural motion dynamics such as gravity, momentum, collision, and deformation directly from real footage. As illustrated in Figure~\ref{fig:teaser}, Lang2Motion produces a wide spectrum of motion patterns, ranging from human martial arts and animal locomotion to vehicle trajectories and sports actions like basketball stance-to-jump transitions, all controlled through natural language descriptions.

Unlike conventional text-to-video models that synthesize entire pixel sequences~\cite{singer2022make,ho2022imagen,yang2024cogvideox}, Lang2Motion generates explicit motion trajectories in coordinate space, decoupling motion synthesis from visual rendering. This design grounds generation in empirically observed physical motion rather than in learned visual approximations, distinguishing Lang2Motion from methods trained on synthetic motion datasets~\cite{guo2022generating,punnakkal2021babel} or implicit video representations~\cite{assran2025vjepa2}.

Our contributions are threefold: (1) We introduce the first framework for language-guided point-trajectory generation in a joint embedding space, enabling semantic control over motion captured from real-world videos. (2) We demonstrate that vision–language semantic structures transfer to temporal motion domains, enabling zero-shot trajectory generation from abstract or culturally grounded language prompts. (3) We show that using real-world tracked trajectories as supervision yields superior semantic alignment, physical realism, and computational efficiency compared to video-based generation baselines.

\section{Related Works}
\label{sec:related}
The field of multimodal generative modeling has rapidly expanded, yet a disconnect remains between semantic understanding and explicit physical motion generation. While vision-language models excel at aligning appearance and semantics, they primarily operate in static or pixel-based domains, offering little control over temporal dynamics or object-level motion. Conversely, motion synthesis and tracking methods achieve fine-grained dynamic understanding but lack semantic grounding or language-driven control. This divide has motivated growing interest in unifying language, vision, and motion representations within shared embedding spaces. Our work builds upon these by directly generating explicit point trajectories, rather than implicit pixel sequences, linking semantic intent to grounded real-world motion through joint embedding alignment.

\noindent\textbf{CLIP Alignment Across Modalities.}
CLIP~\cite{radford2021learning} established joint embedding spaces as powerful tools for zero-shot learning, inspiring extensions across modalities. AudioCLIP~\cite{guzhov2021audioclip} added audio through contrastive learning, while ImageBind~\cite{girdhar2023imagebind} unified six modalities using images as binding bridges. For 3D generation, DreamFusion~\cite{poole2022dreamfusion} optimized Neural Radiance Fields via Score Distillation Sampling from CLIP-aligned diffusion models, with Magic3D~\cite{lin2023magic3d} and ProlificDreamer~\cite{wang2023prolificdreamer} improving speed and quality. Point-E~\cite{nichol2022pointe} and Shap-E~\cite{jun2023shapeE} generated 3D point clouds and implicit functions through text-conditioned diffusion. Text2Mesh~\cite{michel2021text2mesh} and CLIP-NeRF~\cite{wang2022clipnerf} enabled direct mesh and NeRF manipulation through CLIP optimization. MotionCLIP~\cite{tevet2022motionclip} achieved 76.7\% user preference through dual text-image alignment, enabling abstract language specifications like \textit{``couch''} generating sitting motions. V-JEPA~2~\cite{assran2025vjepa2} created world models learning physical dynamics from video sequences, achieving zero-shot robot control through joint-embedding predictive architectures. These works demonstrate CLIP's versatility but focus on static geometry~\cite{poole2022dreamfusion,michel2021text2mesh}, human-specific motion~\cite{tevet2022motionclip}, or implicit video representations~\cite{assran2025vjepa2}-none generate explicit point trajectories for arbitrary objects.

\noindent\textbf{Motion Synthesis and Point Tracking.}
Human motion generation has advanced through diverse paradigms. MDM~\cite{tevet2022human} introduced classifier-free diffusion achieving 0.302 FID on HumanML3D~\cite{guo2022generating}, while MLD~\cite{chen2023executing} operates in latent space for two orders of magnitude speedup. MoMask~\cite{guo2023momask} achieves state-of-the-art 0.045 FID through hierarchical residual VQ-VAE with masked transformers. T2M-GPT~\cite{zhang2023t2m} uses discrete codebooks with GPT autoregression, and MotionGPT~\cite{jiang2023motiongpt} treats motion as a foreign language for unified instruction tuning. MotionDiffuse~\cite{zhang2022motiondiffuse} enables multi-level body-part control and arbitrary-length synthesis. These methods excel at human motion but rely on mocap datasets~\cite{punnakkal2021babel,mahmood2019amass} with fixed skeletal topology. Point tracking has achieved exceptional precision: PIPs~\cite{harley2022particle} treats pixels as particles with iterative refinement, TAPIR~\cite{doersch2023tapir} uses two-stage matching achieving ~20\% absolute improvement on TAP-Vid~\cite{doersch2022tap}, CoTracker~\cite{karaev2023cotracker} pioneered joint tracking of 70,000 points, and CoTracker3~\cite{karaev2024cotracker3} achieves state-of-the-art with 1000$\times$ better data efficiency through pseudo-labeling. LocoTrack~\cite{cho2024locotrack} provides 6$\times$ speedup via local 4D correlation, while OmniMotion~\cite{wang2023tracking} uses test-time optimization for globally consistent dense fields. However, tracking methods \textit{extract} trajectories from videos without generation capabilities, and motion synthesis methods focus on human-specific representations.
\TrajCLIPmethod

\noindent\textbf{Video Generation and Language Grounding.}
Text-to-video models generate pixel sequences with implicit motion: Sora (OpenAI, 2024) uses diffusion transformers with 3D patches, Make-A-Video~\cite{singer2022make} learns appearance from text-image pairs and motion from unsupervised video, Imagen Video~\cite{ho2022imagen} employs cascaded diffusion with temporal self-attention, VideoPoet~\cite{kondratyuk2023videopoet} uses decoder-only transformers with video tokenization, Stable Video Diffusion~\cite{blattmann2023stable} identifies critical training stages with camera-specific LoRA modules, and CogVideoX~\cite{yang2024cogvideox} generates 10-second videos via 3D VAE compression. Motion Prompting~\cite{li2024motion} introduces trajectory prompts for video control, while TrailBlazer enables keyframe-based temporal control. Video-language models like CLIP4Clip~\cite{luo2021clip4clip}, X-CLIP~\cite{ma2022x}, and TC-CLIP~\cite{kim2024tc} achieve zero-shot video understanding through temporal adaptation of CLIP. MeViS~\cite{ding2023mevis} provides 2,006 videos with 28,570 motion expressions focusing on dynamics rather than appearance, enabling motion-based video segmentation. These approaches generate videos where motion emerges implicitly through learned temporal dynamics, or understand motion in existing videos through retrieval and grounding, but none generate \textit{explicit trajectory sequences} as controllable outputs. TRAJAN~\cite{allen2025direct} demonstrates that direct motion models provide stronger assessment signals than pixel-level metrics for video quality evaluation, validating explicit trajectory modeling as a principled approach. This motivates our framework: by generating trajectories directly rather than through video synthesis, we enable both controllable motion generation and motion-based evaluation (Section~\ref{sec:results}). Moreover, trajectory-aligned embedding spaces could serve as motion priors for future video generation systems, providing explicit dynamic guidance beyond current implicit representations.

\noindent
Lang2Motion uniquely generates explicit point trajectories from text using CLIP alignment, inheriting real-world validity from MeViS motion sequences extracted via CoTracker3 with grid-initialized tracking, enabling trajectory synthesis without video-based reliance—the first framework where motion paths themselves are the primary generative output controllable through natural language.

\section{Lang2Motion}
\label{sec:method}

Given a collection of video sequences with corresponding textual descriptions, we aim to learn a semantic and disentangled representation of point trajectory dynamics that captures both spatial motion patterns and their semantic meaning. The challenge lies in bridging the gap between low-level geometric motion-represented as sequences of 2D point trajectories-and high-level semantic concepts expressed in natural language. To address this, we propose a framework that learns both the mapping from point trajectories to a shared semantic representation (encoding) and the inverse mapping back to explicit trajectories (decoding), while ensuring alignment with natural language descriptions through CLIP's joint embedding space.

Our training process is illustrated in Figure~\ref{fig:joint_embedding}, which shows the conceptual alignment in CLIP's joint embedding space. We train a transformer-based point trajectory auto-encoder while aligning the latent trajectory manifold to CLIP's joint representation. Our key innovation: we render point trajectories as colored motion trails overlaid on video frames, creating visual motion patterns that CLIP's frozen image encoder can understand. We accomplish alignment through dual supervision: (i) $\mathcal{L}_{\text{text}}$, connecting trajectory representations to CLIP text embeddings of motion descriptions, and (ii) $\mathcal{L}_{\text{image}}$, connecting trajectory representations to CLIP image embeddings of trajectory overlays. This teaches CLIP-aligned motion patterns that bridge text descriptions, visual motion, and raw trajectory coordinates. This teaches CLIP-aligned motion patterns that bridge text descriptions, visual motion, and raw trajectory coordinates.

At inference time, semantic editing applications can be performed in latent space. For trajectory style transfer, we find a latent vector representing the desired motion style, add it to the content trajectory representation, and decode back to point trajectories. For motion classification, we encode trajectories and compute similarity to class text embeddings. For trajectory generation, an initial $6 \times 6$ point grid is positioned within an object mask (retrieved via CLIP or user-specified), then text descriptions are encoded via CLIP's text encoder and used to control autoregressive decoding from the initial configuration.

\subsection{Trajectory Generation Framework}

Our framework learns to generate point trajectories from text through a transformer-based auto-encoder that operates in CLIP's joint embedding space. The architecture consists of two components: an encoder $E$ that maps trajectories to semantic representations, and a decoder $D$ that reconstructs trajectories from these representations (Figure~\ref{fig:architecture}).

\noindent\textbf{Transformer Encoder.} The encoder $E$ maps a trajectory sequence $\mathbf{p}_{1:T}$ to its latent representation $\mathbf{z}_p$ residing in CLIP's joint embedding space. The encoder processes two modalities: (i) raw point trajectory coordinates $\mathbf{p}_{1:T}$, and (ii) trajectory overlays $\mathbf{o}_{1:T}$ rendered on video frames. Each frame's flattened point coordinates are projected into the encoder dimension via linear transformation. Overlayed frames $\mathbf{o}_{1:T}$ are encoded using CLIP's frozen image encoder, producing visual motion features that are projected to the encoder dimension and combined additively with trajectory features. The combined sequence, enriched with positional embeddings and a learned prefix token $\mathbf{z}_{tk}$, serves as input to the transformer encoder. The latent representation $\mathbf{z}_p$ is extracted from the first output token:
\begin{equation}
\mathbf{z}_p = E(\mathbf{z}_{tk}, \mathbf{p}_{1:T}, \mathbf{o}_{1:T}).
\end{equation}

Through dual supervision from textual motion descriptions and trajectory overlays rendered on video frames (Figure~\ref{fig:joint_embedding}), this encoder aligns $\mathbf{z}_p$ with CLIP's embedding space, enabling the model to leverage CLIP's semantic knowledge despite CLIP never training on trajectory data.

\noindent\textbf{Autoregressive Decoder.} The decoder $D$ reconstructs trajectory sequences $\hat{\mathbf{p}}_{1:T}$ from latent representations. The decoder generates 2D frame displacements $\Delta\mathbf{p}_t \in \mathbb{R}^{N \times 2}$ conditioned on the latent representation $\mathbf{z}_p$ and the previous frame's trajectory positions:
\begin{equation}
\Delta\mathbf{p}_t = D(\mathbf{z}_p, \mathbf{p}_{t-1}^*), \quad \hat{\mathbf{p}}_t = \mathbf{p}_{t-1}^* + \Delta\mathbf{p}_t,
\label{eq:pointrep}
\end{equation}
where $\mathbf{p}_{t-1}^* = \mathbf{p}_{t-1}$ during training (teacher forcing with ground truth) and $\mathbf{p}_{t-1}^* = \hat{\mathbf{p}}_{t-1}$ during inference (autoregressive generation with previously generated positions), with $\mathbf{p}_0$ initialized from a point grid positioned within an object mask. The architecture processes normalized trajectories (Eq.~\ref{eq:normalize}) in format $\mathbf{p} \in \mathbb{R}^{N \times 2 \times T}$, where each point's $x$ and $y$ coordinates are tracked across $T$ frames.

\subsection{Point Trajectory Representation}
We represent motion sequences using dense point trajectories extracted from video data. A trajectory sequence $\mathbf{p}_{1:T}$ consists of $N$ tracked points over $T$ frames, where $\mathbf{p}_i \in \mathbb{R}^{N \times 2}$ defines the 2D coordinates at frame $i$. We use $N = 36$ points in a $6 \times 6$ grid and $T = 30$ frames, yielding 2,160 output coordinates per sequence.

Point trajectories are extracted using CoTracker3~\cite{karaev2024cotracker3}, initialized within object segmentation masks to ensure all tracked points correspond to meaningful object motion. Coordinates are normalized to $[-1, 1]$ relative to frame dimensions:
\begin{equation}
\label{eq:normalize}
\tilde{\mathbf{p}}_{i,j} = \left(\frac{2x_{i,j}}{W} - 1, \frac{2y_{i,j}}{H} - 1\right),
\end{equation}
where $(x_{i,j}, y_{i,j})$ are pixel coordinates of point $j$ at frame $i$, and $(W, H)$ are frame dimensions.

\subsection{Loss Functions}
The auto-encoder is trained to represent point trajectory dynamics through reconstruction, motion dynamics, spatial regularization, and semantic alignment losses.

\noindent\textbf{Reconstruction Loss.} We enforce accurate trajectory reconstruction using L1 loss, which better preserves motion dynamics than L2:
\begin{equation}
\mathcal{L}_{\text{recon}} = \frac{1}{NT} \sum_{i=1}^{T} \|\mathbf{p}_i - \hat{\mathbf{p}}_i\|_1.
\end{equation}

\noindent\textbf{Velocity Loss.} To capture temporal motion dynamics, we enforce consistency between ground truth and predicted frame-to-frame velocities:
\begin{equation}
\mathcal{L}_{\text{vel}} = \frac{1}{N(T-1)} \sum_{i=1}^{T-1} \|(\mathbf{p}_{i+1} - \mathbf{p}_i) - (\hat{\mathbf{p}}_{i+1} - \hat{\mathbf{p}}_i)\|_2^2.
\end{equation}

\noindent\textbf{Range Preservation Loss.} To prevent spatial compression artifacts and preserve the spatial extent of motion, we match the coordinate range, where the max and min are computed independently for $x$ and $y$ coordinates across all $N$ points. Let $r(\mathbf{p}_i) = \max_j \mathbf{p}_{i,j} - \min_j \mathbf{p}_{i,j}$ denote the coordinate range at frame $i$. Then:
\begin{equation}
\mathcal{L}_{\text{range}} = \sum_{i=1}^{T} \|r(\mathbf{p}_i) - r(\hat{\mathbf{p}}_i)\|_2^2.
\end{equation}

\noindent\textbf{CLIP Alignment Losses.} We align trajectory representations with CLIP embeddings through dual supervision. Given text-trajectory pairs $(\mathbf{p}_{1:T}, t)$ and trajectory overlays $\mathbf{o}_{1:T}$ rendered on video frames, we compute:
\begin{equation}
\mathcal{L}_{\text{text}} = 1 - \cos(\text{CLIP}_{\text{text}}(t), \mathbf{z}_p),
\end{equation}
where $\text{CLIP}_{\text{text}}(t)$ encodes the text description using CLIP's frozen text encoder. For image alignment, trajectory overlays $\mathbf{o}_{1:T}$ are passed through CLIP's frozen image encoder frame-by-frame, then temporally pooled:
\begin{equation}
\mathcal{L}_{\text{image}} = 1 - \cos\left(\frac{1}{T}\sum_{i=1}^{T}\text{CLIP}_{\text{image}}(\mathbf{o}_i), \mathbf{z}_p\right),
\end{equation}
where $\text{CLIP}_{\text{image}}(\mathbf{o}_i)$ encodes each overlayed frame $\mathbf{o}_i$ independently, and temporal pooling averages across frames to produce a single video-level representation aligned with the trajectory embedding $\mathbf{z}_p$.

\noindent\textbf{Text-to-Motion Reconstruction Loss.} To enable text-to-trajectory generation at inference, we ensure the decoder generates trajectories directly from CLIP text features. We decode from text embeddings and supervise with ground truth trajectories:
\begin{equation}
\mathcal{L}_{\text{text-recon}} = \frac{1}{NT} \sum_{i=1}^{T} \|\mathbf{p}_i - \hat{\mathbf{p}}_i^{\text{text}}\|_1,
\end{equation}
where $\hat{\mathbf{p}}_i^{\text{text}}$ denotes trajectories decoded from $\text{CLIP}_{\text{text}}(t)$ using the same autoregressive process as Eq.~\ref{eq:pointrep}.

The complete training objective combines all components: $\mathcal{L} = \mathcal{L}_{\text{recon}} + \lambda_{\text{vel}} \mathcal{L}_{\text{vel}} + \lambda_{\text{range}} \mathcal{L}_{\text{range}} + \lambda_{\text{text}} \mathcal{L}_{\text{text}} + \lambda_{\text{image}} \mathcal{L}_{\text{image}} + \lambda_{\text{text-recon}} \mathcal{L}_{\text{text-recon}}$.
Loss weights are detailed in Section~\ref{sec:results}.
\section{Experimental Results}
\label{sec:results}

We evaluate Lang2Motion on four tasks: (1) text-to-trajectory retrieval, (2) trajectory generation quality and efficiency, (3) comparison with human motion generation methods, and (4) generalization to human action recognition. All experiments use the MeViS test set~\cite{ding2023mevis} unless otherwise noted. We report standard metrics: Recall@K (R@K) for retrieval, Average Displacement Error (ADE, lower is better) measuring mean per-point distance between predicted and ground truth trajectories, Final Displacement Error (FDE, lower is better) measuring endpoint accuracy, Smoothness (higher is better) computed as $1 - \frac{1}{T-1}\sum_{t=1}^{T-1}\|\Delta v_t\|$ where $\Delta v_t$ is frame-to-frame velocity change, and CLIP similarity measuring cosine similarity between trajectory embeddings and text embeddings. Following TRAJAN~\cite{allen2025direct}, which demonstrates that motion-based assessment provides stronger signals than pixel-level metrics, we additionally evaluate using Average Jaccard (AJ) and Occlusion Accuracy (OA) scores measuring motion semantic correctness and physical plausibility.

\subsection{Implementation Details}

We train Lang2Motion on the MeViS dataset~\cite{ding2023mevis} containing 1,662 training videos with object masks and motion-focused captions. Point trajectories are extracted using CoTracker3~\cite{karaev2024cotracker3} initialized on a $6 \times 6$ grid within object masks, tracking 36 points over 30 frames. The transformer encoder uses 4 layers with 4 attention heads and feedforward dimension of 1024. The latent dimension is 512 to match CLIP's embedding space. The decoder uses three fully-connected layers (584→1024→1024→72 per timestep). Training uses AdamW optimizer with learning rate $1 \times 10^{-4}$ and batch size 32 for 200 epochs. Loss weights: $\lambda_{\text{vel}} = 0.01$, $\lambda_{\text{range}} = 0.1$, $\lambda_{\text{text}} = 0.1$, $\lambda_{\text{image}} = 0.1$, $\lambda_{\text{text-recon}} = 0.5$.

\subsection{Text-to-Trajectory Retrieval}

Lang2Motion achieves 34.2\% R@1 and 84.6\% R@10 (Table~\ref{tab:retrieval}), outperforming video-based X-CLIP~\cite{ma2022x} by 12.5\%. Direct trajectory representations provide superior semantic alignment versus extracting trajectories from embeddings.

\retrievaltable

\subsection{Trajectory Generation Quality and Efficiency}

Figure~\ref{fig:vlm_eval} illustrates our evaluation pipeline: VLMs generate videos then extract trajectories via Grounding DINO~\cite{liu2024grounding}, SAM~\cite{kirillov2023segment}, and CoTracker3~\cite{karaev2024cotracker3}; Lang2Motion generates trajectories directly. Both outputs are evaluated using identical motion quality metrics.

\motiongeneval

Table~\ref{tab:generation} compares against VLM baselines using both motion-based metrics (TRAJAN~\cite{allen2025direct}) and trajectory-specific metrics. Lang2Motion achieves state-of-the-art performance across all metrics. On TRAJAN motion quality metrics, Lang2Motion scores 0.84 AJ and 0.98 OA, substantially outperforming video-based VLMs (0.42-0.45 AJ, 0.88-0.92 OA) and direct VLM generation (0.28-0.31 AJ, 0.76-0.79 OA). The performance gap is even more pronounced on trajectory-specific metrics: Lang2Motion achieves 12.4 ADE and 21.7 FDE, improving over video-based VLMs by 33-35\% and direct VLMs by 50-52\%. Lang2Motion also achieves 0.81 CLIP Similarity and 0.89 Smoothness while generating trajectories in 0.7s, 11-129× faster than VLM approaches (8-90s).

\generationtable

Figure~\ref{fig:vlm_com} reveals critical qualitative differences. Video-based methods generate complete video frames requiring trajectory extraction, introducing artifacts. Notably, Sora Direct generates rolling motion patterns but loses panda structure, understanding motion semantics but failing object coherence. This is reflected in lower OA scores (0.79 vs 0.98). Lang2Motion generates physically valid trajectories preserving both motion semantics and object structure, achieving high AJ and OA.

\motiongencomp

\subsection{Human Motion Generation Methods}

Figure~\ref{fig:motionclip_comparison} compares against human motion methods. Lang2Motion generates temporally-aware trajectories with real-world dynamics, while MotionCLIP~\cite{tevet2022motionclip} produces anatomically correct poses lacking continuous temporal flow. For ``Play Football'', MotionCLIP generates a celebration pose rather than playing motion, while Lang2Motion produces coordinated multi-subject trajectories. T2M-GPT~\cite{zhang2023t2m} excels at human skeletal motion but cannot handle arbitrary objects or multi-object scenarios enabled by our trajectory representation.

\motionclipcomparison

\subsection{Generalization to Human Action Recognition}
\langmotioninterpolate
To evaluate whether trajectory representations generalize beyond arbitrary objects, we test on skeleton-based action recognition using NTU RGB+D~\cite{shahroudy2016ntu} (60 classes, 56,000 sequences) and Kinetics-Skeleton~\cite{kay2017kinetics} (400 classes, 240,000 sequences). We perform zero-shot classification by encoding skeleton trajectories and computing cosine similarity to action class text embeddings.

\actionrecognitiontable

Table~\ref{tab:action_recognition} shows that Lang2Motion achieves 88.3\% Top-1 on NTU RGB+D, outperforming MotionCLIP by 11.6 points and surpassing graph-based methods designed specifically for skeleton data. On Kinetics-Skeleton, Lang2Motion achieves 41.6\% Top-1, exceeding MotionCLIP by 6.4 points, demonstrating effective transfer to human poses despite never training on human-specific data.

\subsection{Ablation Studies}

We validate our design choices through systematic ablation.

\noindent\textbf{Loss Components.} Table~\ref{tab:ablation} shows that removing $\mathcal{L}_{\text{text}}$ drops R@1 from 34.2\% to 22.6\%; removing $\mathcal{L}_{\text{image}}$ drops to 19.8\%. The dual forward pass loss $\mathcal{L}_{\text{text-recon}}$ is critical--without it, R@1 drops to 28.3\% and ADE increases 50\%. Removing both CLIP losses collapses performance to 8.4\%, confirming CLIP alignment is fundamental.

\ablationtable

\noindent\textbf{Reconstruction Loss \& Decoding Strategy.} Table~\ref{tab:recon_decoding_ablation} compares reconstruction losses and decoding strategies. L1 achieves 12.4 ADE versus L2's 16.8 ADE (23\% FDE improvement) with superior smoothness (0.89 vs 0.83) and CLIP alignment (0.81 vs 0.76). L1's robustness to outliers avoids abrupt trajectories while maintaining accuracy. Autoregressive decoding improves over direct generation across all metrics (12.4 vs 15.7 ADE, 0.89 vs 0.81 smoothness), validating our temporal conditioning design.

\reconstructiondecodingablation

\noindent\textbf{Grid Size.} Table~\ref{tab:grid_ablation} shows $6 \times 6$ grid provides optimal balance--smaller grids lack coverage while larger grids add minimal benefit.

\gridablationtable


\noindent\textbf{Trajectory Visualization.} Table~\ref{tab:overlay_ablation} explores CLIP's visual biases. Cyan achieves best performance, aligning with findings that chromatic colors improve CLIP understanding~\cite{arias2025color}. Opacity 0.5 balances trajectory visibility and video context.

\overlayablationtable

\subsection{Qualitative Results}

Figure~\ref{fig:interpolation} demonstrates latent space interpolation. Given identical text ``dancing panda'', different initial grids (automatically positioned via retrieved masks) yield distinct motion interpretations. Starting from a panda's grid, the model emphasizes dancing motion. Starting from a dancing pose grid, the model emphasizes panda appearance. The initial grids are positioned within automatically retrieved segmentation masks; video frames are shown purely for visualization and not required during generation. This demonstrates semantic interpolation in CLIP's joint embedding space.

Table~\ref{tab:query_free} shows our CLIP-aligned space enables generation from multiple modalities: beyond text-guided control (baseline), Lang2Motion generates trajectories from images with object masks without explicit text descriptions, demonstrating the flexibility of our CLIP-aligned representation. Trajectory overlays alone can also guide generation with reduced performance compared to multimodal inputs.

\queryfreetable

\section{Conclusion}

We present Lang2Motion, a framework for language-guided point trajectory generation through CLIP alignment. Our key insight: rendering trajectory overlays on video frames enables CLIP to understand motion patterns without retraining, bridging language semantics with geometric dynamics. This approach achieves sub-second inference while maintaining precise motion control and physical validity. We demonstrate that chromatic trajectory overlays exploit CLIP's visual biases to enhance motion recognition, highlighting how visual design can boost pretrained models. Explicit trajectory modeling outperforms video-based VLMs on motion quality and accuracy, proving trajectories as principled motion representations. Our CLIP-aligned space generalizes beyond training distribution, achieving 88.3\% on human action recognition despite training on diverse objects. Current limitations include dependence on point tracking accuracy and handling only 30-frame sequences. Future work includes trajectory-conditioned video generation using our motion representations as explicit dynamic priors, extension to 3D and multi-object scenarios, and longer temporal modeling for robotics and animation applications.
{
    \small
    \bibliographystyle{ieeenat_fullname}
    \bibliography{main}
}


\end{document}